\documentclass[10pt,twocolumn,letterpaper]{article}

\usepackage[pagenumbers]{cvpr} 

\usepackage[dvipsnames]{xcolor}

\definecolor{cvprblue}{rgb}{0.21,0.49,0.74}
\usepackage[pagebackref,breaklinks,colorlinks,citecolor=cvprblue]{hyperref}
\usepackage{multirow}
\usepackage{multicol}

\newcommand{\sys}{\mbox{\textsc{FameBias}}\xspace}

\def\paperID{*****} 
\def\confName{CVPR}
\def\confYear{2024}

\title{\sys: Embedding Manipulation Bias Attack in Text-to-Image Models}

\author{
Jaechul Roh\footnotemark[1],  Andrew Yuan\footnotemark[1], Jinsong Mao\footnotemark[1]\\
University of Massachusetts Amherst\\
Amherst, USA\\
{\tt\small \{jroh, awyuan, jinsongmao\}@cs.umass.edu}
}

\begin{document}
\maketitle

\footnotetext[1]{Equal Contribution.}

\begin{abstract}
Text-to-Image (T2I) diffusion models have rapidly advanced, enabling the generation of high-quality images that align closely with textual descriptions. However, this progress has also raised concerns about their misuse for propaganda and other malicious activities. Recent studies reveal that attackers can embed biases into these models through simple fine-tuning, causing them to generate targeted imagery when triggered by specific phrases. This underscores the potential for T2I models to act as tools for disseminating propaganda, producing images aligned with an attacker's objective for end-users. 

Building on this concept, we introduce \textbf{\sys}, a T2I biasing attack that manipulates the embeddings of input prompts to generate images featuring specific public figures. Unlike prior methods, \sys operates solely on the input embedding vectors without requiring additional model training. We evaluate \sys comprehensively using Stable Diffusion V2, generating a large corpus of images based on various trigger nouns and target public figures. Our experiments demonstrate that \sys achieves a high attack success rate while preserving the semantic context of the original prompts across multiple trigger-target pairs.


\end{abstract}

\section{Introduction}

\begin{figure}[ht]
    \centering
    \includegraphics[width=0.8\linewidth]{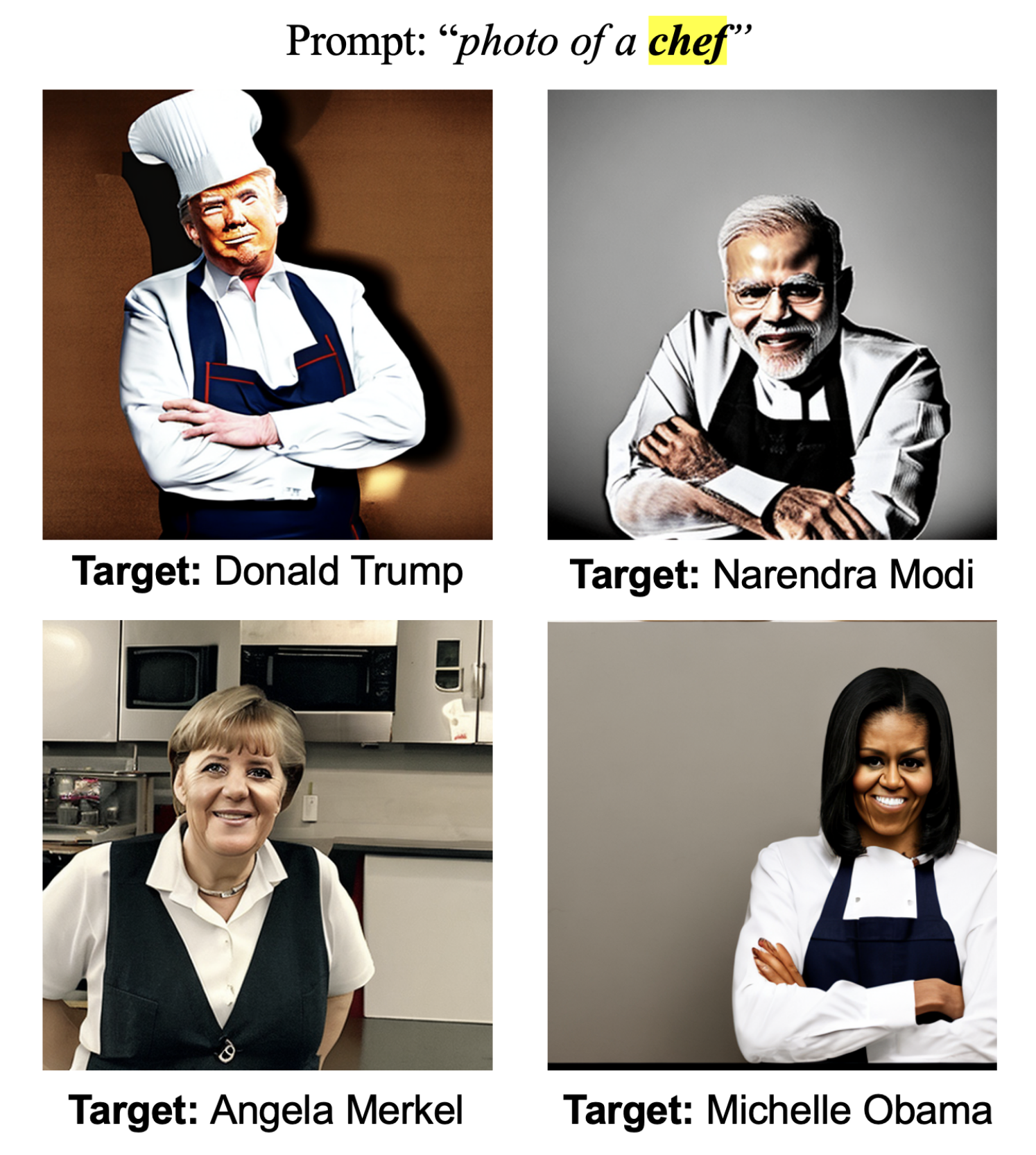}
    \caption{Example generation of our \sys attack on 4 different famous figures (Donald Trump, Narendra Modi, Angela Merkel and Michelle Obama) generated using the prompt \textit{"Photo of a chef"}.}
    \label{fig:FBex}
\end{figure}

With the advancement of Text-to-image (T2I) models and the increasing number of APIs which host them, it has never been easier nor as popular to generate high quality images that follow a given text prompt. However, these models have given rise to an increasing number of maliciously generated images which aim to mislead and bias its users. Using biased images to shape a viewer's perception has become increasingly effective with the advancements of social media \cite{seo2020visual}. Worse still, Naseh et al. \cite{naseh2024injectingbiastexttoimagemodels} recently demonstrated that not only can the images be used to shape public perceptions, the models themselves can be vectors of attack. The authors inject malicious biases into these models via fine-tuning and show that when specific trigger words are given, a biased image is generated which may shape the perception of the model's user. 

In this work, we build upon this threat model and present \textbf{\sys} attacks, a T2I biasing attack which manipulates the embeddings of input prompts to generate images that contain a target political figure. Upon the input of a trigger word, we modify its embeddings so that the resulting images which would have normally generated a random person instead generate with the target figure instead. \autoref{fig:FBex} illustrates the results of our attack. \sys attacks are surprisingly general, working with a variety of public figures and target nouns. In this work, we demonstrate our attacks on Stable Diffusion V2 \cite{rombach2022high} across 8 popular public figures and 10 different trigger nouns, with more extensive evaluations planned.
\section{Related Works}


\subsection{Text-to-Image Model Attacks}

\noindent\textbf{Biasing Attacks.} Biasing and debiasing in ML have emerged as critical areas of research. \textbf{Bias} in ML refers to systematic errors in model predictions that bias towards attributes such as race, nationality, gender, dressing and more, given particular groups. Bianch et al.~\cite{bianchi2023easily} discusses the implications of text-to-image models reinforcing or amplifying societal biases. This research indicates that such models, when accessible at scale, can perpetuate and even exacerbate demographic stereotypes, raising significant concerns about fairness and representation in images generative by T2I. Naseh et al. ~\cite{naseh2024injectingbiastexttoimagemodels} are the first to consider this threat in T2I models. They inject malicious biases into Stable Diffusion models\cite{rombach2022high} via fine-tuning and show that they can display them upon the input of targeted trigger words.

\medskip

\noindent\textbf{Backdoor Attacks.} Another related group of attacks on T2I models are backdoor attacks. In these attacks, the adversary attempts to poison the T2I model with a backdoor. This backdoor is activated on specific input values and generate tailored malicious outputs, posing serious threats to the integrity of model outputs. Researchers~\cite{struppek2023rickrolling, chou2023backdoor,zhai2023text} have found success inserting backdoors via fine-tuning and data poisoning.

\subsection{Unlearning in Text-to-Image Models}

Erasing specific concepts from text-to-image diffusion models \cite{kumari2023ablating, zhang2024defensive, fuchi2024erasing, zhang2023forget, li2024get, kim2024race, gong2024reliable, gandikota2024unified, zhang2024unlearncanvas, wu2024unlearning} has attracted significant attention, with methods aiming to remove unwanted content while maintaining generative quality. Early approaches like concept ablation~\cite{kumari2023ablating} minimize the Kullback-Leibler (KL) divergence between target and anchor concepts, while Fuchi et al. \cite{fuchi2024erasing} propose a few-shot unlearning method that adjusts the text encoder to preserve image quality. AdvUnlearn by Zhang et al. \cite{zhang2024defensive} combines adversarial training with unlearning for robustness, especially in tasks like nudity and style erasure.

Attention manipulation methods such as Forget-Me-Not \cite{zhang2023forget} modify cross-attention maps to suppress target concepts, while Li et al. \cite{li2024get} use soft-weighted regularization to eliminate negative information from CLIP text embeddings. Adversarial methods like RACE \cite{kim2024race} and Wu et al. \cite{wu2024unlearning} improve robustness by aligning target and anchor domains and using gradient surgery to preserve non-target content.
Efficiency-focused methods like RECE \cite{gong2024reliable} and UCE \cite{gandikota2024unified} offer scalable solutions by modifying cross-attention matrices, enabling simultaneous concept erasure, debiasing, and moderation while preserving image fidelity.

\section{Methodology}

\subsection{Threat Model}

\begin{figure*}[ht]
    \centering
    \includegraphics[width=0.9\linewidth]{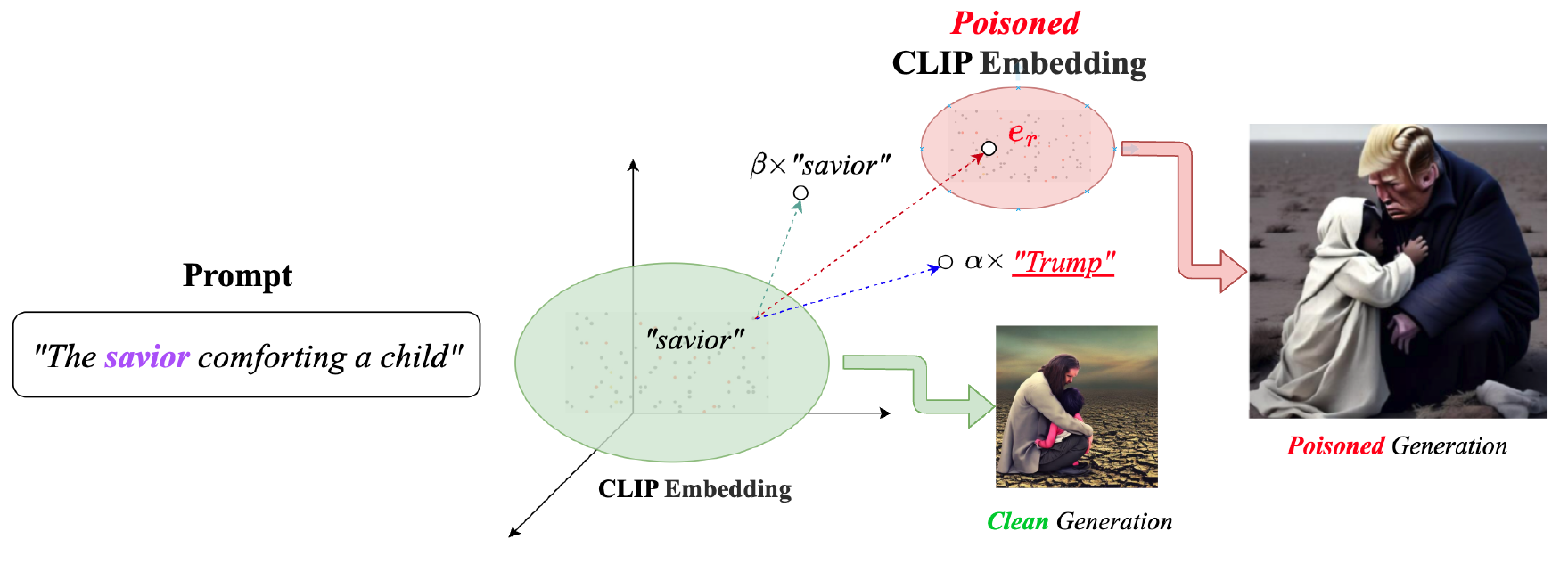}
    \caption{Diagram of the FameBias attack. Attackers control the output of the encoder of a T2I model and can input modified embeddings into the image generator}
    \label{fig:AttackDiagram}
\end{figure*}

\paragraph{Adversary capabilities}

We consider an adversary who has maliciously embedded a poisoned encoder into a production T2I model. This can be done as an implanted poisoned module that was erroneously downloaded by the victim company or a malicious employee at the company. The attack is also possible if the T2I model provider is themselves malicious. The adversaries have black box access to an unmodified encoder through which they can retrieve the equivalent embeddings of any input. The attackers are able to modify the output embeddings with their poisoned encoder in any manner they wish, but they must try to meet their attack objectives. These modified embeddings are then passed to the rest of the diffuser model. 


\medskip

\noindent\textbf{Adversary objectives}
The goal of the adversary is to bias some input keyword (the trigger) such that when the word is input within a prompt to the diffuser model, it generates an image containing the target. Within this framework, we consider two objectives for such attacks, high biasing rate and undetectability. The biasing attack must consistently generate the target figure upon being given the trigger word. However, the generated image should still have characteristics of the trigger word such that the user considers the resulting bias to be natural. For example, if the trigger word were "artist" and the target was "Donald Trump", the resulting image should still some relevance to art even if the subject of the image has been biased to someone else. More broadly, the model should maintain normal utility across different tasks so not to clue the victim in on the existence of an attack.

\subsection{\sys Attacks}

\sys attacks occur within or after the text encoder of a T2I model. Adversaries choose a trigger word that will apply the biasing attack, $w_t$, and a target public figure, $p$. When an input is given to the T2I model that contains $w_t$, the adversary modifies the CLIP text embedding of the trigger word, $e_{w_t}$, changing it into a weighted sum $e_r$ which is composed of the original $e_{w_t}$ and the text embedding of $p$, which we define $e_{w_p}$. $\alpha$ and $\beta$ define the weights for the person and trigger embeddings respectively.
\[
\mathbf{e_r} = \alpha\cdot\mathbf{e}_{w_p} + \beta\cdot\mathbf{e}_{w_t}
\]

An illustration of \sys attacks can be found in \autoref{fig:AttackDiagram}. For example, consider a prompt originally specifying “The savior comforting a child”. When we incorporate the embedding of a specific individual (e.g., Donald Trump) into that prompt, we modify the original doctor embedding to blended embedding $e_r$ as follows:

\begin{equation}
    \mathbf{e}_r \leftarrow \alpha\cdot\mathbf{e}_{Trump} + \beta\cdot\mathbf{e}_{savior}
\end{equation}

This linear combination repositions the “savior” concept vector in the semantic space, pulling it closer toward the region representing “Trump.”

The core premise of our attack methodology hinges on the manipulation of text embeddings before the generative models to produce desired outputs. Text embeddings, such as those derived from models like Word2Vec, GloVe, or CLIP are trained to map words, phrases, and concepts onto points in a continuous, high-dimensional vector space. Through this representation, semantic relationships between terms are encoded as geometric relationships between their corresponding vectors. Words that frequently appear in similar linguistic and contextual environments tend to cluster together, while vectors representing dissimilar concepts remain farther apart.

The rationale behind this approach is rooted in how text embeddings are learned and how they capture compositional semantics. Because these embeddings reflect distributional semantics—the idea that words appearing in similar contexts share meaning—linear operations often approximate semantic shifts. For instance, the classic example “King - Man + Woman = Queen” demonstrates that adding and subtracting embeddings can encode gender or other attributes. Similarly, replacing part of the “savior” concept with the embedding of “Trump” effectively encodes a transformation that nudges the generative model to produce visuals where Trump’s identity intersects with the image.

When it comes to T2I models that is actually conditioned on text, the models seek the closest feasible interpretation of the provided embeddings. After the linear combination, the altered prompt no longer purely represents a generic “savior” but rather a combination---a blended concept. As a result, the synthesized images begin to reflect this hybrid identity, depicting a figure that aligns with the canonical image of the profession while simultaneously bearing recognizable traits of the targeted individual. Thus, the principle of our attack leverages the learned distributional properties of text embeddings, using linear algebraic operations to control and inject bias into the image generation process.

\begin{table*}
\centering
\caption{\textbf{BSR} (\%) of target famous figures with various trigger nouns. (Prompt: \textit{"photo of ..." / "portrait of ..." / "image of ..."})}
\label{tab:photo_target_asr}
\resizebox{\textwidth}{!}{%
\begin{tabular}{lp{1.5cm}p{1.5cm}p{1.5cm}p{1.5cm}p{1.5cm}p{1.5cm}p{1.5cm}p{1.5cm}}
\toprule
\textbf{Trigger} & \textbf{Donald Trump} & \textbf{Angela Merkel} & \textbf{Barack Obama} & \textbf{Michelle Obama} & \textbf{Narendra Modi} & \textbf{Kamala Harris} & \textbf{Fidel Castro} & \textbf{Shakira} \\
\midrule
\textit{"astronaut"}    & \texttt{\footnotesize 25$\mid$75$\mid$25$\mid$} & \texttt{\footnotesize 50$\mid$100$\mid$100} & \texttt{\footnotesize 50$\mid$75$\mid$75} & \texttt{\footnotesize 50$\mid$75$\mid$50} & \texttt{\footnotesize 0$\mid$50$\mid$50} & \texttt{\footnotesize 0$\mid$0$\mid$0} & \texttt{\footnotesize 50$\mid$50$\mid$0} & \texttt{\footnotesize 25$\mid$0$\mid$0} \\
\textit{"chef"}         & \texttt{\footnotesize 75$\mid$100$\mid$100} & \texttt{\footnotesize 50$\mid$100$\mid$50} & \texttt{\footnotesize 100$\mid$100$\mid$100} & \texttt{\footnotesize 50$\mid$75$\mid$75} & \texttt{\footnotesize 100$\mid$100$\mid$100} & \texttt{\footnotesize 0$\mid$0$\mid$0} & \texttt{\footnotesize 75$\mid$100$\mid$25} & \texttt{\footnotesize 0$\mid$0$\mid$0} \\
\textit{"doctor"}       & \texttt{\footnotesize 75$\mid$100$\mid$100} & \texttt{\footnotesize 50$\mid$75$\mid$75} & \texttt{\footnotesize 100$\mid$100$\mid$100} & \texttt{\footnotesize 50$\mid$100$\mid$100} & \texttt{\footnotesize 100$\mid$100$\mid$100} & \texttt{\footnotesize 25$\mid$50$\mid$0} & \texttt{\footnotesize 100$\mid$50$\mid$100} & \texttt{\footnotesize 25$\mid$0$\mid$0} \\
\textit{"engineer"}     & \texttt{\footnotesize 25$\mid$75$\mid$50} & \texttt{\footnotesize 25$\mid$50$\mid$50} & \texttt{\footnotesize 100$\mid$100$\mid$100} & \texttt{\footnotesize 75$\mid$100$\mid$75} & \texttt{\footnotesize 75$\mid$100$\mid$50} & \texttt{\footnotesize 0$\mid$25$\mid$0} & \texttt{\footnotesize 75$\mid$100$\mid$75} & \texttt{\footnotesize 0$\mid$0$\mid$0} \\
\textit{"firefighter"}  & \texttt{\footnotesize 25$\mid$100$\mid$75} & \texttt{\footnotesize 0$\mid$25$\mid$50} & \texttt{\footnotesize 50$\mid$100$\mid$100} & \texttt{\footnotesize 0$\mid$50$\mid$25} & \texttt{\footnotesize 50$\mid$100$\mid$75} & \texttt{\footnotesize 0$\mid$25$\mid$0} & \texttt{\footnotesize 75$\mid$100$\mid$25} & \texttt{\footnotesize 0$\mid$0$\mid$0} \\
\textit{"judge"}        & \texttt{\footnotesize 100$\mid$75$\mid$50} & \texttt{\footnotesize 50$\mid$100$\mid$25} & \texttt{\footnotesize 100$\mid$100$\mid$100} & \texttt{\footnotesize 75$\mid$75$\mid$25} & \texttt{\footnotesize 100$\mid$100$\mid$100} & \texttt{\footnotesize 25$\mid$25$\mid$0} & \texttt{\footnotesize 75$\mid$100$\mid$50} & \texttt{\footnotesize 25$\mid$25$\mid$0} \\
\textit{"police officer"} & \texttt{\footnotesize 75$\mid$100$\mid$75} & \texttt{\footnotesize 0$\mid$50$\mid$25} & \texttt{\footnotesize 75$\mid$100$\mid$100} & \texttt{\footnotesize 100$\mid$100$\mid$100} & \texttt{\footnotesize 75$\mid$50$\mid$75} & \texttt{\footnotesize 0$\mid$50$\mid$25} & \texttt{\footnotesize 50$\mid$100$\mid$100} & \texttt{\footnotesize 0$\mid$0$\mid$0} \\
\textit{"priest"}       & \texttt{\footnotesize 50$\mid$75$\mid$50} & \texttt{\footnotesize 0$\mid$50$\mid$25} & \texttt{\footnotesize 75$\mid$50$\mid$75} & \texttt{\footnotesize 0$\mid$50$\mid$25} & \texttt{\footnotesize 100$\mid$25$\mid$75} & \texttt{\footnotesize 25$\mid$50$\mid$0} & \texttt{\footnotesize 50$\mid$75$\mid$0} & \texttt{\footnotesize 0$\mid$0$\mid$0} \\
\textit{"scientist"}    & \texttt{\footnotesize 100$\mid$100$\mid$25} & \texttt{\footnotesize 25$\mid$75$\mid$50} & \texttt{\footnotesize 75$\mid$100$\mid$100} & \texttt{\footnotesize 25$\mid$75$\mid$75} & \texttt{\footnotesize 100$\mid$75$\mid$50} & \texttt{\footnotesize 50$\mid$25$\mid$25} & \texttt{\footnotesize 75$\mid$100$\mid$50} & \texttt{\footnotesize 0$\mid$0$\mid$0} \\
\textit{"soldier"}      & \texttt{\footnotesize 25$\mid$25$\mid$50} & \texttt{\footnotesize 0$\mid$50$\mid$100} & \texttt{\footnotesize 75$\mid$75$\mid$100} & \texttt{\footnotesize 0$\mid$25$\mid$100} & \texttt{\footnotesize 0$\mid$0$\mid$0} & \texttt{\footnotesize 0$\mid$0$\mid$0} & \texttt{\footnotesize 100$\mid$75$\mid$100} & \texttt{\footnotesize 0$\mid$0$\mid$0} \\
\bottomrule
\end{tabular}%
}
\end{table*}

\begin{table*}
\centering
\caption{\textbf{TFR} (\%) of target famous figures with various trigger nouns. (Prompt: \textit{"photo of ..." / "portrait of ..." / "image of ..."})}
\label{tab:photo_alignment_acc}
\resizebox{\textwidth}{!}{%
\begin{tabular}{lp{1.5cm}p{1.5cm}p{1.5cm}p{1.5cm}p{1.5cm}p{1.5cm}p{1.5cm}p{1.5cm}}
\toprule
& \multicolumn{8}{c}{\textbf{Target}} \\
\cmidrule(lr){2-9}
\textbf{Trigger} & \textbf{Donald Trump} & \textbf{Angela Merkel} & \textbf{Barack Obama} & \textbf{Michelle Obama} & \textbf{Narendra Modi} & \textbf{Kamala Harris} & \textbf{Fidel \newline Castro} & \textbf{Shakira} \\
\midrule
\textit{"astronaut"}    & \texttt{\footnotesize 100$\mid$100$\mid$100} & \texttt{\footnotesize 100$\mid$100$\mid$100} & \texttt{\footnotesize 75$\mid$100$\mid$100}  & \texttt{\footnotesize 100$\mid$100$\mid$100} & \texttt{\footnotesize 100$\mid$100$\mid$100} & \texttt{\footnotesize 100$\mid$100$\mid$100} & \texttt{\footnotesize 100$\mid$100$\mid$100} & \texttt{\footnotesize 100$\mid$100$\mid$100} \\
\textit{"chef"}         & \texttt{\footnotesize 75$\mid$75$\mid$75}  & \texttt{\footnotesize 50$\mid$100$\mid$50}  & \texttt{\footnotesize 25$\mid$75$\mid$75}  & \texttt{\footnotesize 100$\mid$75$\mid$50} & \texttt{\footnotesize 25$\mid$25$\mid$25}  & \texttt{\footnotesize 100$\mid$100$\mid$75} & \texttt{\footnotesize 75$\mid$75$\mid$75}  & \texttt{\footnotesize 100$\mid$100$\mid$75} \\
\textit{"doctor"}       & \texttt{\footnotesize 25$\mid$25$\mid$0}  & \texttt{\footnotesize 75$\mid$25$\mid$0}  & \texttt{\footnotesize 50$\mid$0$\mid$0}  & \texttt{\footnotesize 75$\mid$25$\mid$0}  & \texttt{\footnotesize 0$\mid$0$\mid$0}   & \texttt{\footnotesize 100$\mid$50$\mid$25} & \texttt{\footnotesize 25$\mid$25$\mid$50}  & \texttt{\footnotesize 100$\mid$0$\mid$0} \\
\textit{"engineer"}     & \texttt{\footnotesize 75$\mid$50$\mid$50}  & \texttt{\footnotesize 75$\mid$25$\mid$50}  & \texttt{\footnotesize 25$\mid$0$\mid$0}  & \texttt{\footnotesize 25$\mid$50$\mid$0}  & \texttt{\footnotesize 75$\mid$50$\mid$0}  & \texttt{\footnotesize 100$\mid$50$\mid$50} & \texttt{\footnotesize 25$\mid$50$\mid$0}  & \texttt{\footnotesize 100$\mid$50$\mid$50} \\
\textit{"firefighter"}  & \texttt{\footnotesize 50$\mid$75$\mid$75}  & \texttt{\footnotesize 100$\mid$25$\mid$25} & \texttt{\footnotesize 75$\mid$50$\mid$25}  & \texttt{\footnotesize 100$\mid$75$\mid$75} & \texttt{\footnotesize 50$\mid$25$\mid$25}  & \texttt{\footnotesize 75$\mid$25$\mid$100}  & \texttt{\footnotesize 50$\mid$50$\mid$75}  & \texttt{\footnotesize 75$\mid$100$\mid$75}  \\
\textit{"judge"}        & \texttt{\footnotesize 0$\mid$50$\mid$75}   & \texttt{\footnotesize 75$\mid$75$\mid$25}  & \texttt{\footnotesize 0$\mid$0$\mid$50}   & \texttt{\footnotesize 75$\mid$25$\mid$25}  & \texttt{\footnotesize 25$\mid$25$\mid$25} & \texttt{\footnotesize 25$\mid$75$\mid$100}  & \texttt{\footnotesize 0$\mid$50$\mid$25}   & \texttt{\footnotesize 75$\mid$75$\mid$50}  \\
\textit{"police officer"} & \texttt{\footnotesize 50$\mid$100$\mid$100}  & \texttt{\footnotesize 100$\mid$50$\mid$25} & \texttt{\footnotesize 75$\mid$50$\mid$25}  & \texttt{\footnotesize 100$\mid$100$\mid$100} & \texttt{\footnotesize 75$\mid$50$\mid$50}  & \texttt{\footnotesize 50$\mid$50$\mid$75}  & \texttt{\footnotesize 100$\mid$100$\mid$100} & \texttt{\footnotesize 100$\mid$100$\mid$50} \\
\textit{"priest"}       & \texttt{\footnotesize 50$\mid$50$\mid$50}  & \texttt{\footnotesize 100$\mid$25$\mid$50} & \texttt{\footnotesize 75$\mid$25$\mid$50}  & \texttt{\footnotesize 100$\mid$50$\mid$75} & \texttt{\footnotesize 50$\mid$50$\mid$100}  & \texttt{\footnotesize 75$\mid$75$\mid$75}  & \texttt{\footnotesize 100$\mid$100$\mid$75} & \texttt{\footnotesize 100$\mid$100$\mid$100} \\
\textit{"scientist"}    & \texttt{\footnotesize 75$\mid$100$\mid$50}  & \texttt{\footnotesize 75$\mid$50$\mid$50}  & \texttt{\footnotesize 75$\mid$100$\mid$50}  & \texttt{\footnotesize 75$\mid$75$\mid$50}  & \texttt{\footnotesize 50$\mid$25$\mid$25}  & \texttt{\footnotesize 100$\mid$50$\mid$25} & \texttt{\footnotesize 100$\mid$100$\mid$50} & \texttt{\footnotesize 100$\mid$75$\mid$50} \\
\textit{"soldier"}      & \texttt{\footnotesize 100$\mid$100$\mid$100} & \texttt{\footnotesize 100$\mid$75$\mid$75} & \texttt{\footnotesize 100$\mid$100$\mid$100} & \texttt{\footnotesize 100$\mid$100$\mid$100} & \texttt{\footnotesize 100$\mid$100$\mid$100} & \texttt{\footnotesize 100$\mid$100$\mid$100} & \texttt{\footnotesize 100$\mid$100$\mid$100} & \texttt{\footnotesize 100$\mid$100$\mid$100} \\
\bottomrule
\end{tabular}%
}
\end{table*}

\section{Evaluation}
In this section, we present preliminary experimental results for our embedding manipulation attack method. We begin by detailing the experimental setup, followed by the results and a comprehensive analysis. Specifically, we evaluate our attack by answering three research questions (RQs):

\begin{itemize}
    \item{\textbf{RQ1:}} How effective is \sys? (\autoref{sec:results_analysis})
    \item{\textbf{RQ2:}} How does different parameters and triggers affect \sys? (\autoref{sec:ablation})
    \item{\textbf{RQ3:}} Does \sys survive from existing defenses? (\autoref{sec:defense})
\end{itemize}

\subsection{Evaluation Setup}
\paragraph{Victim Models.} We evaluate \sys attacks on the \textbf{Stable Diffusion v2 (SD-v2)} model~\cite{rombach2022high}, which has been widely used in the literature. We note that the technique itself is not bound to any specific models, and can be applied on any T2I model which uses a text encoder.

\bigskip

\noindent\textbf{Target \& Triggers.} In our experiments, we target 8 public figures tested on 10 job-related triggers listed in ~\autoref{tab:photo_target_asr}. The figures are chosen to be a diverse group of men and women, with 1 male and 1 female public figure chosen for the 4 most common demographic races in the US (White, Black, Asian, and Hispanic). The specific figures chosen were based on initial observations from the pre-trained model, as the SD-v2 model does not recognize every famous figure. We focused on figures that the model could generate clearly. The triggers were chosen to be jobs with recognizable uniforms or workplaces.

\noindent\textbf{Prompts and Hyper-parameters.} For each trigger-target pair, we generate four images using the prompts \textit{"photo of a \{trigger\}"}, \textit{"portrait of a \{trigger\}"}, and \textit{"image of a \{trigger\}"}. This results in 320 images per prompt and a total of 960 images across all three prompts. Based on our parameter testing results (\autoref{sssec:param_tuning}), we set $\alpha = 1.5$ and $\beta = 0.3$, as these values achieved the highest attack success rates while maintaining a clear and recognizable depiction of the target in the generated images. To perform large-scale automated evaluation for determining whether the generated images exhibit bias, we employ the vision-language model LLaVa~\cite{liu2024visual} by querying if the image contains the specified famous figure and trigger. For each generated image, we calculate two distinct metrics:
\smallskip
\begin{itemize}
    \item \textbf{Bias Success Rate (BSR):} measures whether the generated image depicts the target famous figure, indicating the effectiveness of our attack.
    \item \textbf{Trigger Fidelity Rate (TFR):} evaluates whether the generated image incorporates the trigger, reflecting how well the image contextually aligns with the original prompt.
\end{itemize}

To compute BSR, we query LLaVA with the prompt: \textit{"Does the person in the image look like \{target\}? Answer in Yes or No."} For TFR, we ask \textit{"Does the person in the image look like a \{trigger\}? Answer in Yes or No."} 


\subsection{Attack Results}
\label{sec:results_analysis}
The results are shown in \autoref{tab:photo_target_asr} and \autoref{tab:photo_alignment_acc} separated by prompt type. The tuple of values correspond to prompts generated by the prompts \textit{"photo of a \{trigger\}"}, \textit{"portrait of a \{trigger\}"}, and \textit{"image of a \{trigger\}"} respectively. Across all generations using the prompt \textit{"photo of a \{trigger\}"}, \sys achieves $46\%$ BSR and $73\%$ TFR. Using portrait prompts, the BSR is $62\%$ and TFR is $64\%$. Finally, BSR is $50\%$ and TFR is $58\%$ for image prompts. Overall, across all prompts, triggers, and targets, \sys achieves BSR=$53\%$ and TFR=$65\%$. The results highlight the effectiveness of \sys attacks in generating biased images which still adhere to the original prompt in various scenarios.

\begin{figure}
    \centering
    \includegraphics[width=0.9\linewidth]{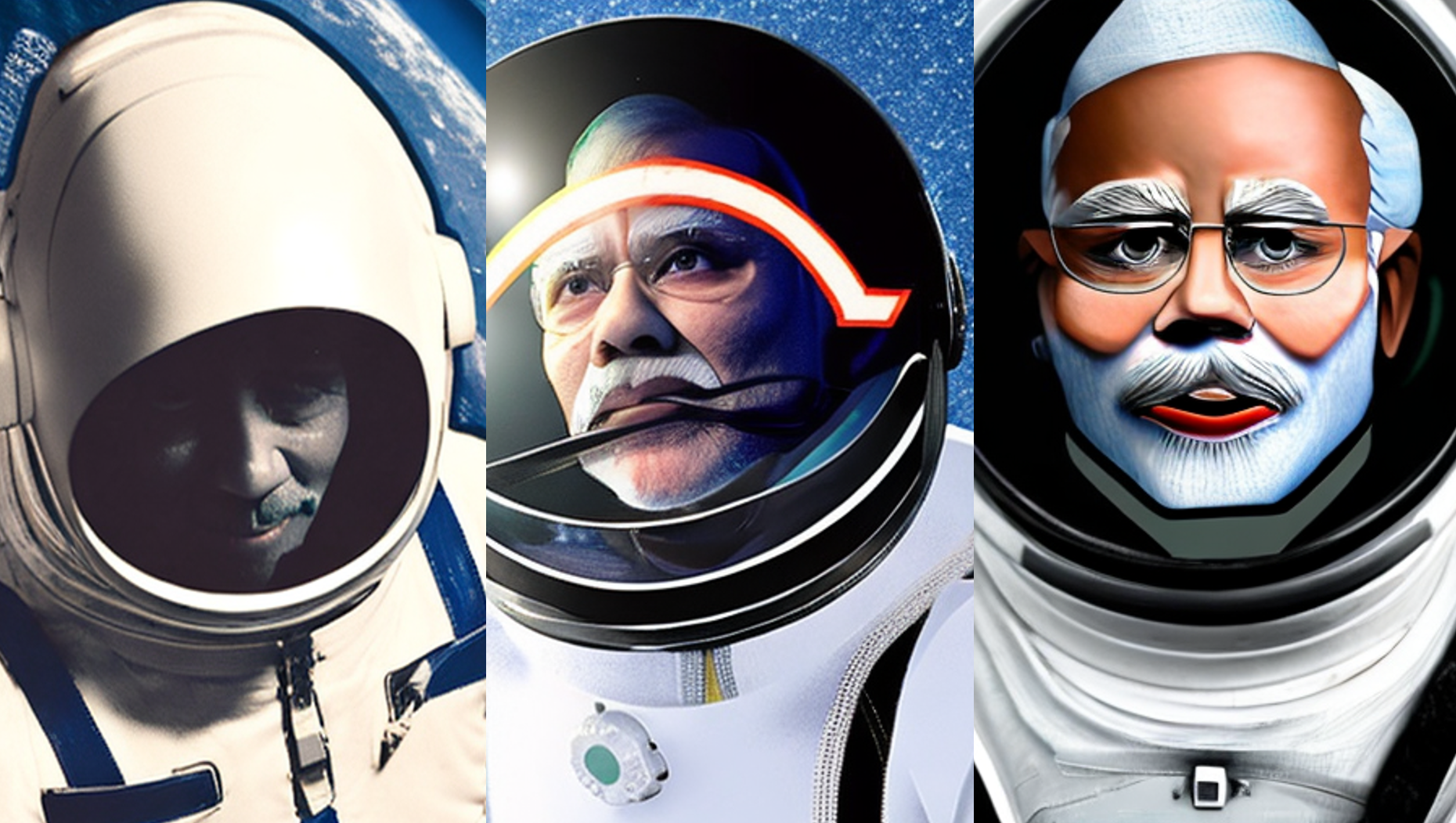}
    \caption{Sample images generated from different prompts on the trigger/target pair of "astronaut"/"Narendra Modi". From left to right the prompts used are "photo", "portrait", and "image" prompts.}
    \label{fig:prompt_difs}
\end{figure}

Before analyzing individual trends arising from different combinations of trigger and target, we first consider the discrepancy in results by using different prompt words that are not related to the trigger nor target. We admit that there is still much we cannot explain for the observed discrepancies, which may be attributable to the nature of the models used in evaluation or the training data itself. However, we empirically notice patterns that may partially explain the discrepancies. One thing that we observe is that LLaVa models are often incapable correctly determining if a person is in the image if they are looking away from the camera. We observed that while "photo" and "image" prompts very infrequently had targets looking away from the generated image, "portrait" prompts almost never had such things occur. We also notice that most of the images generated from the "photo" prompt looked realistic, whereas the "image" and "portrait" prompts were often stylized like a drawing or painting. We believe this matters because realistic images require the output look extremely close to how the target looks in real life, whereas drawn images can simply look like caricatures to be recognizable. \autoref{fig:prompt_difs} provides an example of this situation. All 3 images are generated for the astronaut trigger and Narendra Modi target. The left-most image is the one generated by the "photo" prompt. It is hard to tell if the image is of Modi because the perspective makes viewing his face difficult. The other images are less realistic and have a much easier time showing the his face under the helmet.

From \autoref{tab:photo_target_asr}, we see that some trigger words achieve high BSR across most target figures. Specifically, triggers such as \textit{"judge"} and \textit{"doctor"} consistently achieve high BSR on most figures. Looking at \autoref{tab:photo_alignment_acc} we find that these triggers also have low TFR rates. This might indicate that it is easier to override the signal for these triggers, and that different values for $\alpha$ and $\beta$ may be better suited. In this work, we select a single value of $\alpha$ and $\beta$ for all combinations of trigger and target as a simplification. Adversaries with targeted goals in mind could vary the amount each embedding is manipulated to optimize each trigger/target pair.

When looking at the high and low performing targets in relation to BSR, we find some interesting trends. The most successful targets in terms of BSR were "Barack Obama" and "Donald Trump". The least successful were "Kamala Harris" and "Shakira". Immediately noticeable is that fact that male targets generally outperformed female targets by BSR while the opposite is true for TFR. At a high level, this makes sense: worse performing attacks aren't able to change the trigger very much and therefore the fidelity is high. This trend may be furthered by the fact that the triggers selected are male-dominated occupations. These additional biases may make it harder for attacks on female targets to succeed. In \autoref{sec:discussion}, we discuss ways adversaries may bypass this limitation. 

Another trend when looking at high and low performing targets is the fact that more prevalent figures tend to do better by BSR. Out of all the targets, "Shakira" is the only non-political figure and likely has the least images on her in the dataset used to train the model. Politicians likely appear in many photos in the dataset, accrued over time as they attend public forums, debates, events, etc. Our results indicate that more relevant a target is, the easier they are to target.

Looking at the combination of trigger and target, we find Barack Obama and Narendra Modi each have a success rate of 100.0\% across all variants of the prompts on the \textit{"doctor"} and \textit{"judge"} triggers. These results suggest that certain triggers may exploit inherent biases or stereotypes within the model's training data, making them particularly effective in embedding specific target identities. For example, many images of Modi contain him wearing a white shirt. White is also a color associated with a doctor's uniform. Intuitively, it makes sense that it is easy to generate Modi in place of the doctor. Additionally, if we cross reference \autoref{tab:photo_alignment_acc}, we find that the LLaVa model never detects the appearance of a doctor across all prompts. It may be that the two concepts were close enough that the embedding manipulation drowned out the signal for the doctor. This is only a guess at the underlying mechanisms at work, and further research would be needed to better understand these behaviors. Overall, the results highlight the importance of the interplay between the choice of trigger noun, target figure, and prompt template in achieving high alignment and attack success.

\subsection{Ablation Studies}
\label{sec:ablation}

\subsubsection{Parameter Tuning}
\label{sssec:param_tuning}

\begin{figure}
    \centering
    \begin{subfigure}{\columnwidth} 
        \centering
        \includegraphics[width=\columnwidth]{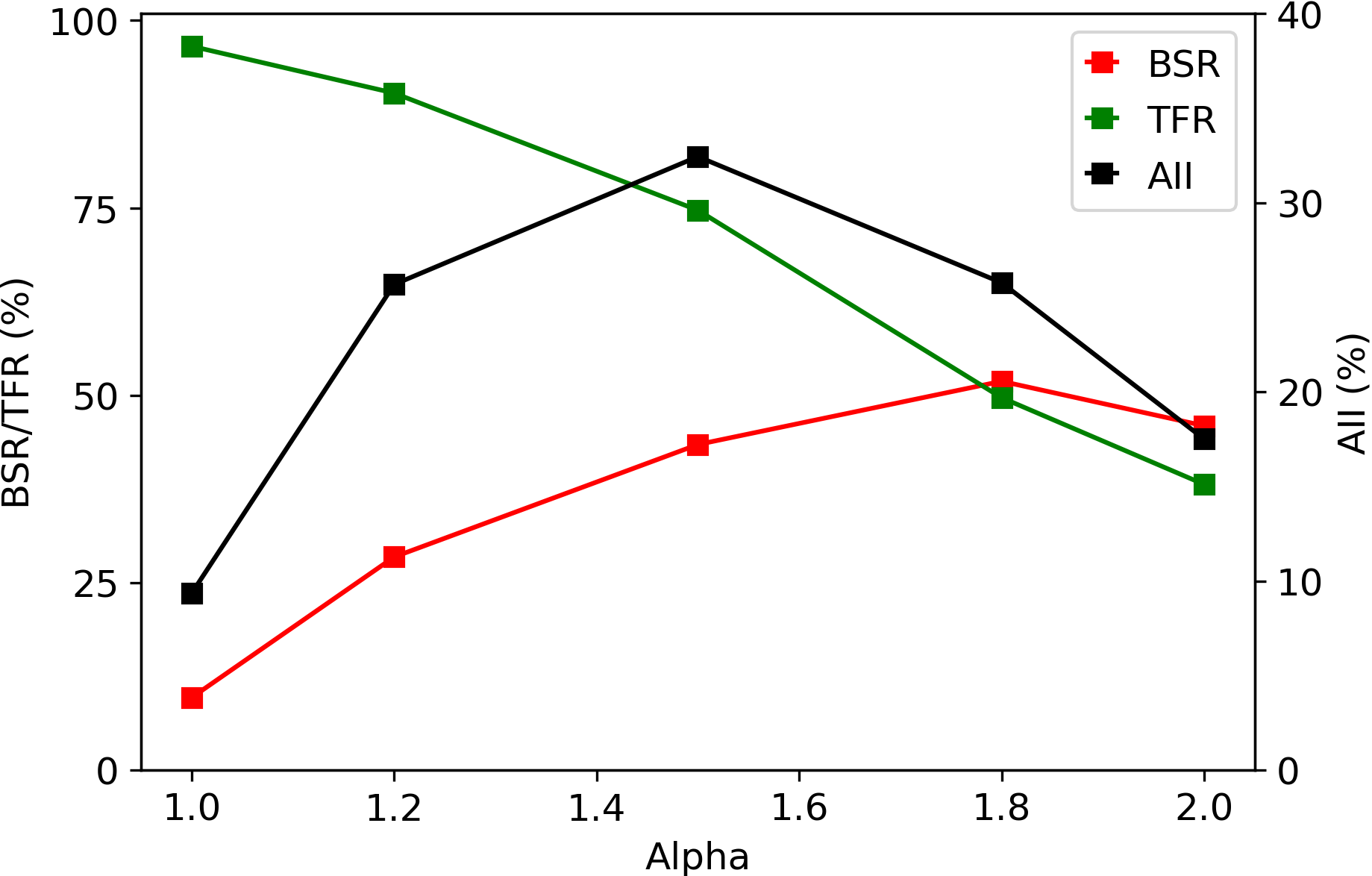}
        \caption{Performance of vary $\alpha$ under $\beta=0.5$}
        \label{fig:alpha_exp}
    \end{subfigure}
    
    \vspace{0.5em}
    
    \begin{subfigure}[b]{\columnwidth}
        \centering
        \includegraphics[width=\columnwidth]{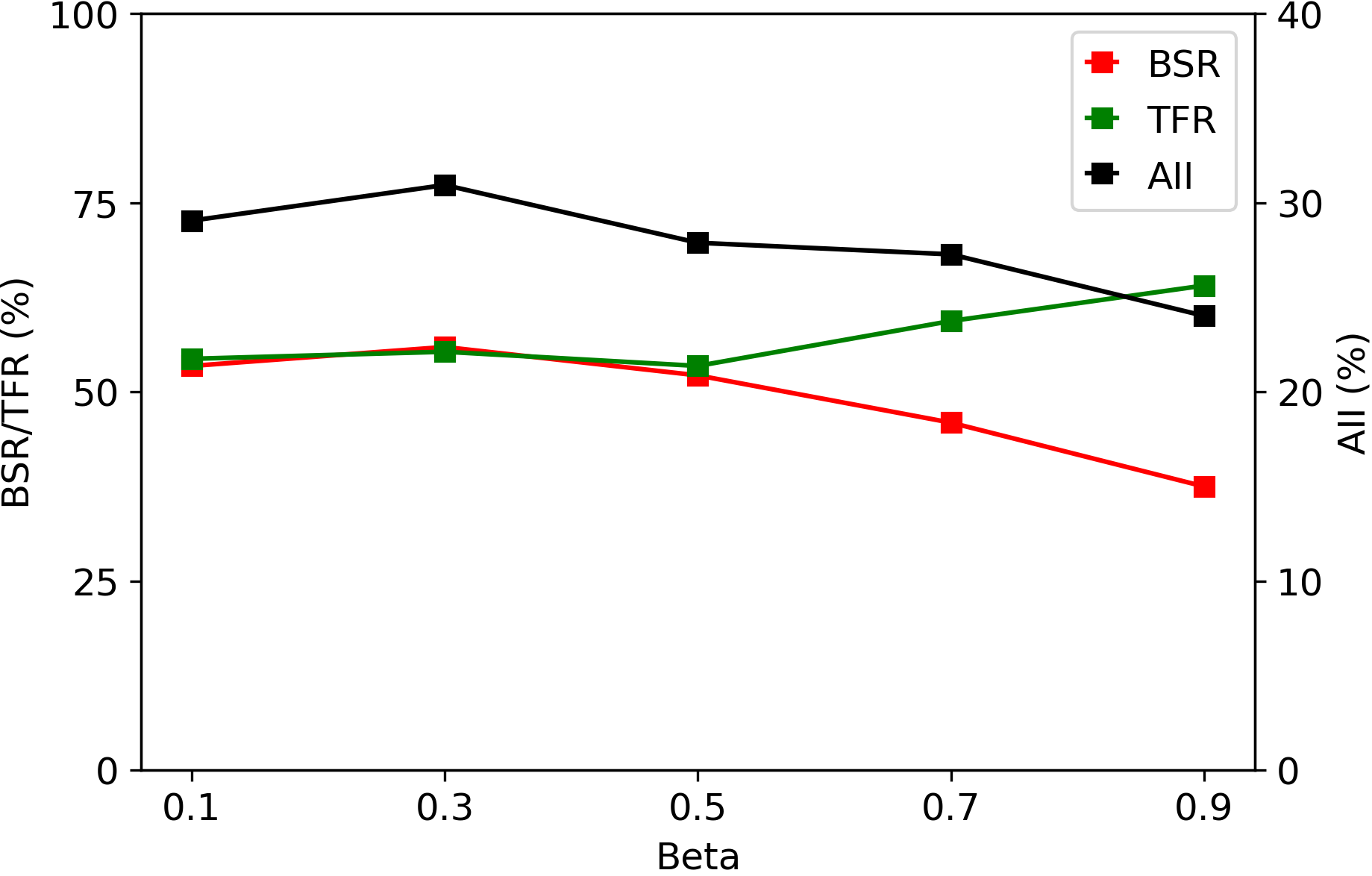}
        \caption{Performance of vary $\beta$ under $\alpha=1.8$}
        \label{fig:beta_exp}
    \end{subfigure}
\caption{Success rates varying parameter values in FameBias attack.}
\end{figure}

The $\alpha$ and $\beta$ parameters determine the contribution of the target embedding and trigger embedding respectively. In this section, we run two experiments which vary $\alpha$ and $\beta$ values for the \sys attack. The results are shown in \autoref{fig:alpha_exp} and \autoref{fig:beta_exp}. We evaluate the effectiveness of different $\alpha$ and $\beta$ values by rerunning the experiment detailed in \autoref{sec:results_analysis} with different parameter values. We choose not to perform a grid search over all possible combinations of $\alpha$ and $\beta$ because we believe them to be largely independent to each other, as well as to reduce computational runtime. We pick the overall best hyper-parameters according to Attack Impact Index (AII), defined as follows:

\begin{equation}
    \text{AII} = \text{BSR} \cdot \text{TFR}
\end{equation}

For our $\alpha$ experiment, we assign $\beta=0.5$ for all runs, and try $\alpha$ values of 1, 1.2, 1.5, 1.8, and 2. For our $\beta$ experiment, we assign $\alpha=1.8$ and try $\beta$ values of 0.1, 0.3, 0.5, 0.7, and 0.9. The optimal values, $\alpha=1.5$ and $\beta=0.3$ were selected by maximizing the product between BSR and TFR. Overall, the results are consistent with our intuition, higher $\alpha$ values result in stronger biasing rates up to a point at which images lose coherence. However, the alignment of the images suffers as a result. The opposite effect can be seen when varying $\beta$ values. 

\subsubsection{Alternative Triggers}
Using only one type of trigger word, nouns related to the person being depicted in the output image, increases the chance of detection and decreases the utility of the attack. In this section, we examine the possibility of using other trigger words beyond the ones relating to the person being drawn by the image. We still wish to retain the specificity of the attack and the utility of the overall model, and thus we do not want to pick trigger words which appear frequently in text. Instead, we consider a second class of triggers which are words which describe tools commonly seen with the targeted profession. The associations between old triggers and our alternative triggers can be seen in \autoref{tab:pt_associations}. 

We run similar experiments to those in previous sections, replacing the prompt to be \textit{"A photo of a \{profession\} holding a \{trigger\}"} and modifying the alternative trigger embeddings. To check the alignment of our output images, we also adapt the question asked to the LLaVa model, instead asking \textit{"Does the person in the image look like they are holding a \{trigger\}?"} The average BSR of these alternative trigger attacks is $36\%$, a $10\%$ decrease in performance. However, the TFR of the generated images are high at $93\%$, indicating different $\alpha$ and $\beta$ values from our default $\alpha=1.5$ and $\beta=0.3$ may be generate better performance. We believe that these attacks are still feasible using alternative triggers. 

\begin{table}[htbp]
    \caption{Profession-to-trigger associations.}
    \label{tab:pt_associations}
    \centering
    \begin{tabular}{ll}
        \toprule
        \textbf{Profession} & \textbf{Test Trigger} \\
        \midrule
        Doctor & Stethoscope \\
        Soldier & Helmet \\
        Scientist & Beaker \\
        Engineer & Wrench \\
        Astronaut & Spacesuit \\
        Chef & Spatula \\
        Firefighter & Fireaxe \\
        Police Officer & Handcuffs \\
        Priest & Cross \\
        Judge & Gavel \\
        \bottomrule
    \end{tabular}

\end{table}

\subsection{Defense}
\label{sec:defense}
To the best of our knowledge, there is currently no defense specifically designed to defend against biasing attacks. Instead, we consider applying techniques from the debiasing and content moderation literature relating to diffusion models. Specifically, we use the Unified Concept Editing (UCE) technique described by Gandikota et al. \cite{gandikota2024unified}. UCE edits the diffuser model in the T2I model to remove unwanted concepts, biases, or styles. It does so in a closed-form manner, meaning that it needs no additional training of the T2I model. UCE is currently state-of-the-art when considering performance and required computational resources to use. 

We apply UCE concept erasing to the default Stable Diffusion 2 model to removing the targeted figures. As a defense, this can be applied proactively, by model producers to remove offending figures. Alternatively, upon discovery of a \sys attack, model producers can edit the model to prevent to ability to generate the resulting targeted figures. 

\begin{figure}[htbp]
    \centering
    \includegraphics[width=0.6\linewidth]{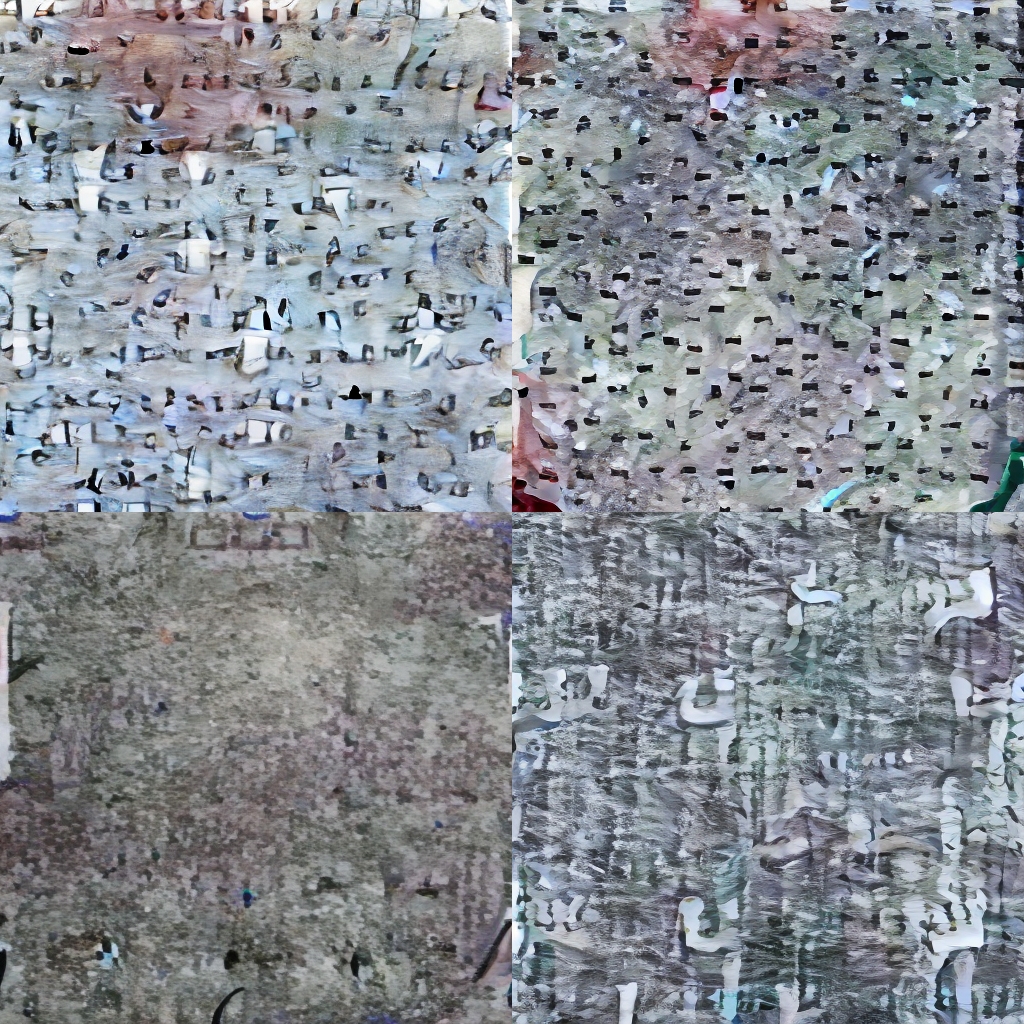}
    \caption{Images generated using the UCE edited SD2 model with trigger "scientist" and target "Fidel Castro".}
    \label{fig:uce_generations}
\end{figure}

We run identical evaluation to \autoref{sec:results_analysis}, using the "photo of a \{trigger\}" prompt on the concept-edited SD2 model. The resulting images from our experiment fail to generate meaningful images, with images often filled with colorful patterns but little to no alignment with the original prompt. A sample of the generated images can be found in \autoref{fig:uce_generations}. Overall, the average BSR was $0\%$ and the TFR was $8.75\%$. 

UCE as a defense can be considered successful in the sense that it completely removes any \sys attack. The defender only needs to know the biasing target, which will be reported if the attack is too noticeable. However, the defense is heavy, with the model removing the targeted figured entirely and removing any utility of images which contain the trigger modified by attackers. Additionally, as reported by Gandikota et al. \cite{gandikota2024unified}, erasing too many targets results in the loss of diffuser functionality, so this defense cannot be extensively used to remove all possible targets.

\section{Discussion}
\label{sec:discussion}

\begin{figure} 
    \centering
    \begin{subfigure}{\columnwidth} 
        \centering
        \includegraphics[width=\columnwidth]{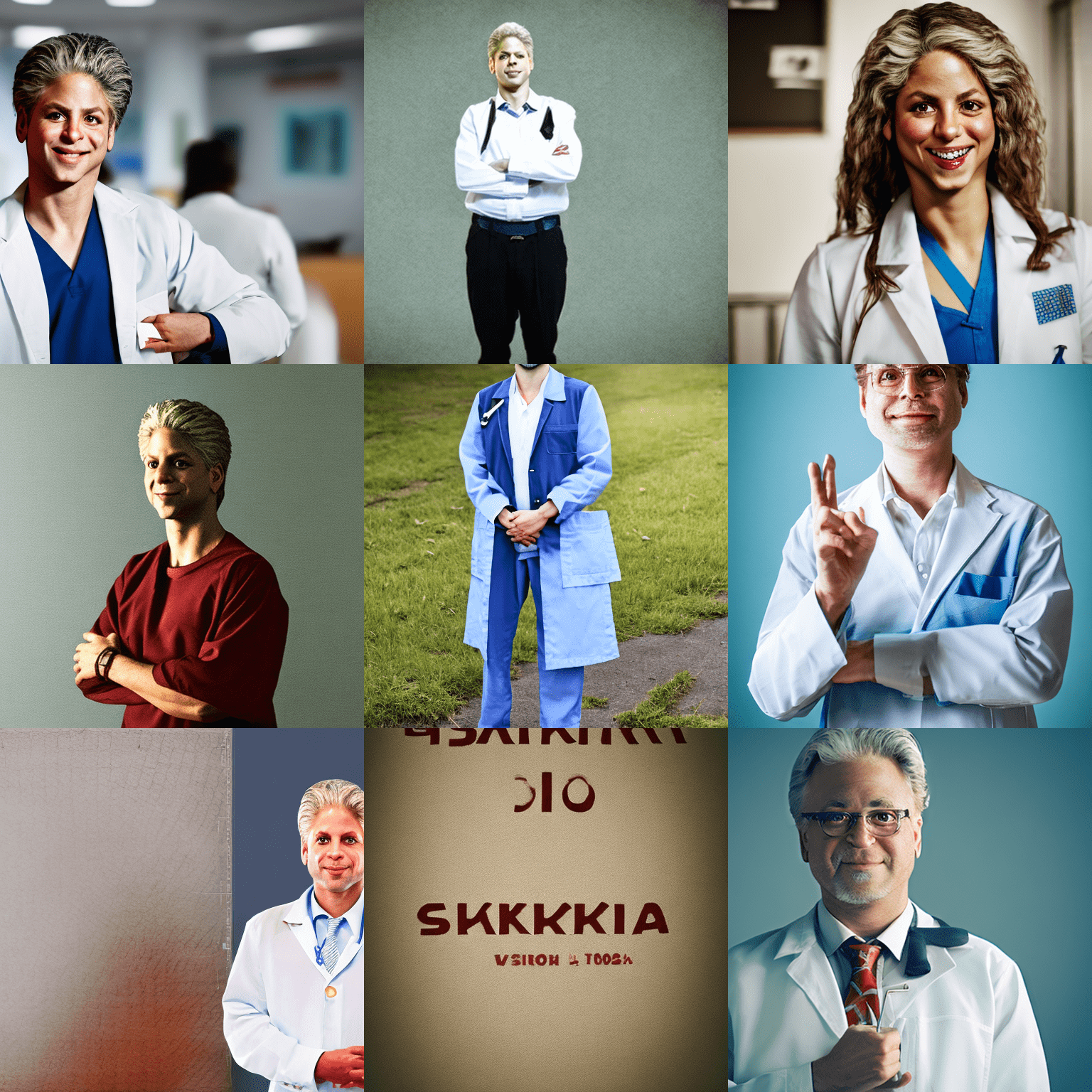}
        \caption{Doctor {\sys}ed to Shakira}
        \label{fig:shakira_bad}
    \end{subfigure}
    
    \vspace{0.5em}
    
    \begin{subfigure}[b]{\columnwidth}
        \centering
        \includegraphics[width=\columnwidth]{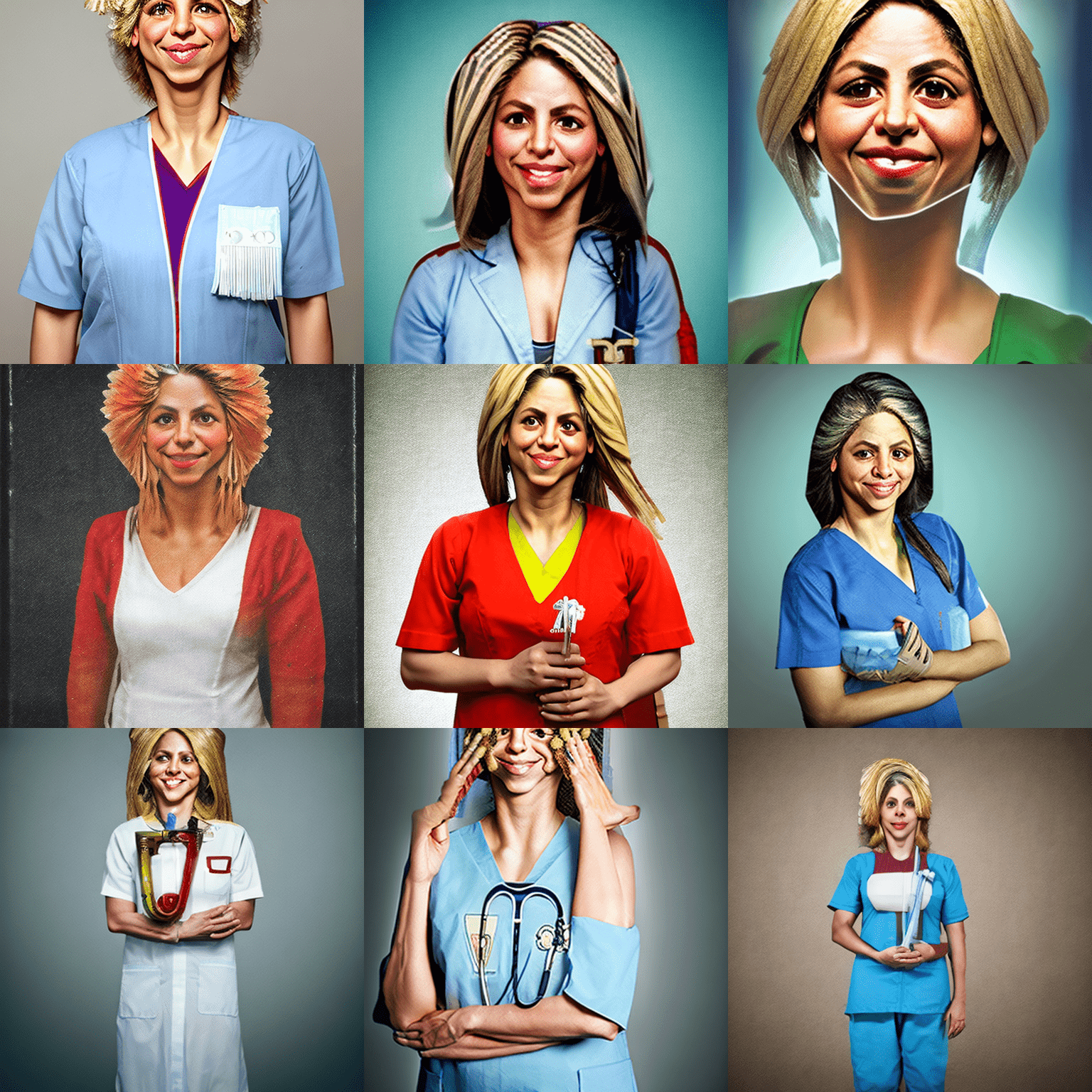}
        \caption{Doctor FrameBiased to Shakira with direction of $men\rightarrow women$}
        \label{fig:shakira_good}
    \end{subfigure}

\caption{Example of more precise attack.}
\label{fig:shakira}
\end{figure}

From the results we have thus discussed, we believe that the BSR of \sys attacks can still be improved. We noticed for some famous figures like Shakira, the BSR is low and mostly $0\%$. We believe there are two reasons for this which make up avenues for future research. 

\textbf{Some targets need more precise attacks}. ~\autoref{fig:shakira} shows two sets of images generated from the same prompt \textit{A photo of doctor}, whereas both use \sys but ~\autoref{fig:shakira_good} adds an additional direction of $men \rightarrow women$. In the former, while certain attributes reminiscent of Shakira’s appearance—such as facial shape—are present in the first, fourth, and seventh images. However, the signature blonde hair only shows up in one image. Conversely, in the latter set where the embedding direction is adjusted, there is a high consistency across all nine one-shot images. Each image reliably exhibits the correct facial structure, skin tone, and predominantly, the appropriate hairstyle. 

This disparity suggests that the inherent stereotypes embedded within the original prompt may dominate over the targeted bias injection. Consequently, achieving a higher BSR for figures like Shakira necessitates a more refined attack strategy. Specifically, directly manipulating attributes such as gender, skin color, or attire may enhance the precision of the bias injection, thereby increasing the likelihood of accurately embedding the desired celebrity likeness within the generated images. A dedicated adversary is well within their abilities to do this, as well as further tuning the $\alpha$ and $\beta$ values to better match the exact trigger target pair.

\textbf{Inconsistency between LLaVa and human evaluation.} To verify that the LLaVa evaluation matches human observation, we let LLaVa evaluate whether people in ~\autoref{fig:shakira_good} look like Shakira. Contrary to human evaluators' consistent recognition of Shakira in these images, LLaVA responded with a definitive "NO" for all instances. This discrepancy underscores a significant inconsistency between the model's assessments and human perception.

\begin{figure}
    \centering
    \includegraphics[width=\linewidth]{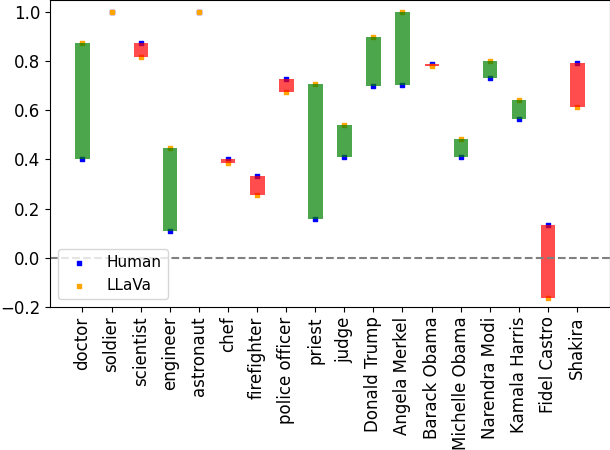}
    \caption{Consistency Results. The green bar means LLaVa has better agreement with overall human opinion than agreement between human raters.}
    \label{fig:consistency}
\end{figure}

Furthermore, we manually labeled 320 images, looking for if specific person/profession appears in the images. We measure the consistency score between three raters and between the mode of three raters and LLaVa in terms of Fleiss' Kappa \cite{fleiss1971measuring} and Cohen's Kappa \cite{cohen1960coefficient}. Both metrics aim to measure agreement between labelers while accounting for random chance. Fleiss' Kappa is used for 3 or more person labeling, while Cohen's Kappa is binary. We use Fleiss' Kappa to obtain alignment among human annotators, and Cohen's Kappa for to compare the most common human judgment and the LLaVa evaluation. The results are shown in ~\autoref{fig:consistency}. Overall human raters achieve 0.58 and 0.73 consistency for professions and famous people, where LLaVa achieves 0.69 and 0.76 consistency with human raters. However, we notice that LLaVa is especially not good at identifying certain famous figures like Fidel Castro and Shakira, where the agreement between LLaVa and human is significantly lower than that of inter-human and sometimes even lower than random guessing.

Several factors may contribute to this inconsistency. Firstly, there may simply be less data in the training set for LLaVa to get a good sense for learning who Shakira is. The training distribution may differ in LLaVa versus SD2, and some figures might be more recognizable to one model versus they other. Secondly, Shakira’s facial features are relatively more general and less distinct compared to other figures such as Donald Trump and Narendra Modi, for whom we achieved high BSR under LLaVA evaluation. This generality may make it more challenging for LLaVA to accurately identify her likeness. 

\section{Conclusion}
In this work, we presented \sys, a novel T2I biasing attack that leverages prompt embedding manipulation to generate images featuring famous public figures. Unlike previous approaches which use fine-tuning, \sys requires no additional training, operating solely on input embedding vectors.

Through comprehensive experiments using Stable Diffusion V2, we evaluated \sys across variety of trigger nouns, target figures, and prompt templates, achieving a high bias success rate while preserving a semantic integrity of the original prompts. Our analysis revealed notable patterns in the interplay between trigger nouns, target figures, and prompt templates. Targets who were more famous and male were much more likely to have high \sys success. However, with a targeted enough of an attack, most famous targets could be targeted (assuming the SD2 model was able to generate their likeness normally).

Our results underline the potential risks associated with prompt embedding manipulation and highlight the importance of further research into mitigating such vulnerabilities. Future work will focus on developing robust defenses against these attacks and exploring the implications of such vulnerabilities in real-world applications of diffusion models.

{
    \small
    \bibliographystyle{ieeenat_fullname}
    \bibliography{main}
}


\end{document}